\documentclass[11pt]{article}

\usepackage{graphicx}%
\usepackage{multirow}%
\usepackage{amsmath,amssymb,amsfonts}%
\usepackage{amsthm}%
\usepackage{mathrsfs}%
\usepackage[title]{appendix}%
\usepackage{xcolor}%
\usepackage{textcomp}%
\usepackage{manyfoot}%
\usepackage{booktabs}%
\usepackage{algorithm}%
\usepackage{algorithmicx}%
\usepackage{algpseudocode}%
\usepackage{listings}%
\usepackage{pdfpages}
\usepackage[letterpaper,top=2cm,bottom=2cm,left=3cm,right=3cm,marginparwidth=1.75cm]{geometry}
\usepackage{bigstrut}
\usepackage{xspace}
\usepackage{longtable}
\usepackage{caption}
\usepackage{subcaption}
\usepackage{adjustbox}
\usepackage[colorlinks=true, linkcolor=black, citecolor=blue]{hyperref}
\usepackage[superscript]{cite}
\usepackage{lineno}
\usepackage{enumitem}
\usepackage{xcolor}

\ifdefined\highlighton
    \definecolor{customdefaultcolor}{RGB}{18, 159, 87} 

    \newcommand{\highlight}[2][customdefaultcolor]{\textcolor{#1}{#2}}
\else
    \newcommand{\highlight}[2][customdefaultcolor]{#2}
\fi

\makeatletter

\renewcommand{\maketitle}{\bgroup\setlength{\parindent}{0pt}
\begin{flushleft}
  \textbf{\@title}

  \@author
\end{flushleft}\egroup
}
\makeatother

\begin{document}

\newcommand{\method}{MedAISys\xspace}


\title{A Perspective for Adapting Generalist AI to Specialized Medical AI Applications and Their Challenges}
\author{Zifeng Wang$^{1}$, Hanyin Wang$^{1,2}$, Benjamin Danek$^1$,  Ying Li$^3$, Christina Mack$^{5}$, Luk Arbuckle$^{5}$, Devyani Biswal$^{5}$, 
Hoifung Poon$^{6}$, Yajuan Wang$^7$, Pranav Rajpurkar$^8$, \\Jimeng Sun$^{1,4\#}$\\
$^1$ Department of Computer Science, University of Illinois Urbana-Champaign, Champaign, IL \\
$^2$  Division of Hospital Internal Medicine, Mayo Clinic Health System,  Mankato, MN\\
$^3$ Regeneron, Tarrytown, NY \\
$^4$ Carle Illinois College of Medicine, University of Illinois Urbana-Champaign, Champaign, IL \\
$^5$ IQVIA, Durham, NC  \\
$^6$ Microsoft Research, Redmond, WA  \\
$^7$  Teladoc Health, Purchase, NY  \\
$^8$ Department of Biomedical Informatics, Harvard Medical School, Boston, MA \\
$^\#$ Corresponding author. Email: jimeng@illinois.edu}

\maketitle



\abstract{
The integration of Large Language Models (LLMs) into medical applications has sparked widespread interest across the healthcare industry, from drug discovery and development to clinical decision support, assisting telemedicine, medical devices, and healthcare insurance applications. This perspective paper aims to discuss the inner workings of building LLM-powered medical AI applications and introduces a comprehensive framework for their development. We review existing literature and outline the unique challenges of applying LLMs in specialized medical contexts. Additionally, we introduce a three-step framework to organize medical LLM research activities: 1) {\it Modeling:} breaking down complex medical workflows into manageable steps for developing medical-specific models; 2) {\it Optimization:} optimizing the model performance with crafted prompts and integrating external knowledge and tools, and 3) {\it System engineering:} decomposing complex tasks into subtasks and leveraging human expertise for building medical AI applications. Furthermore, we offer a detailed use case playbook that describes various LLM-powered medical AI applications, such as optimizing clinical trial design, enhancing clinical decision support, and advancing medical imaging analysis. 
Finally, we discuss various challenges and considerations for building medical AI applications with LLMs, such as handling hallucination issues, data ownership and compliance, privacy, intellectual property considerations, compute cost, sustainability issues, and responsible AI requirements. 
}

\newpage

\section{Introduction}\label{sec:intro}
Artificial intelligence (AI) is increasingly being integrated into various medical tasks, including clinical risk prediction~\cite{choi2016retain}, medical image understanding~\cite{wang2022medclip}, and synthetic patient records generation~\cite{wang2022promptehr}. Typically, these models are designed for specific tasks and will struggle with unfamiliar tasks or out-of-distribution data~\cite{moor2023foundation}. Large language models (LLMs) are foundation models characterized by their extensive training data and enormous model scale. Unlike traditional AI models, LLMs demonstrate an emergent capability in language understanding and the ability to tackle new tasks through in-context learning~\cite{brown2020language}. For example, we can teach LLMs to conduct a new task by providing the text explanation of the task (or ``prompts"), the input and output protocols, and several examples. This adaptability has sparked interest in employing generalist LLMs to medical AI applications such as chatbots for outpatient reception~\cite{wan2024outpatient}. Contrary to the common belief that generalist LLMs will excel in many fields~\cite{Yang2024-vh}, we advocate that domain-specific AI adaptations for medicine are more effective and safe. This paper will overview how various adaptation strategies can be developed for medical AI applications and the associated trade-offs.

\begin{figure}[htbp]
    \centering
    \includegraphics[width=\linewidth]{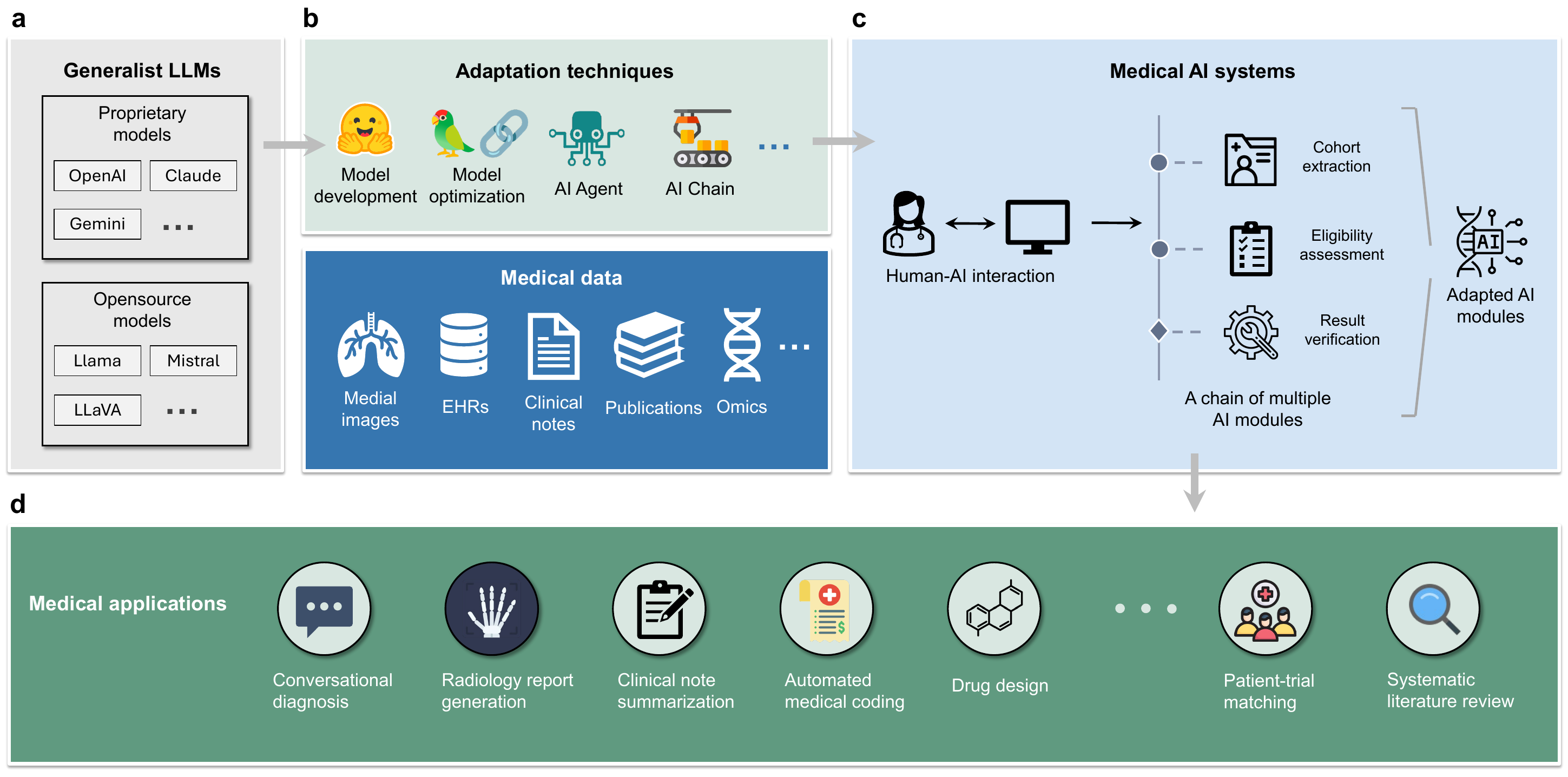}
    \caption{Workflow for adapting generalist LLMs for medical AI through adaptation techniques. (a) Generalist AI models, such as proprietary systems (e.g., Open AI's GPT-4) and open-source models (e.g., LLaMA), serve as foundational technologies for developing specialized medical AI models.
(b) Adapting generalist AI to medical tasks involves several techniques, including model fine-tuning, prompt optimization, and the development of AI agents or AI chains. These methods use diverse medical datasets, such as medical images, electronic health records (EHRs), clinical notes, publications, and omics data, to enhance AI model training and performance.
(c) Effective system engineering for medical AI entails integrating AI modules into comprehensive chains to support tasks like cohort extraction, eligibility assessment, and result verification. This process emphasizes human interaction and AI, resulting in tailored AI modules for specific applications.
(d) Generalist AI applications in medicine span various domains, including conversational diagnosis, radiology report generation, clinical note summarization, automated medical coding, drug design, patient-trial matching, and systematic literature reviews. All require advanced system integration for optimal performance.
}
\label{fig:overview}
\end{figure}

Generalist LLMs such as ChatGPT can support broad tasks but may underperform in specialized domains~\cite{tian2024opportunities}. One notable drawback is the occurrence of ``hallucinations," which are fabricated facts that look plausible yet incorrect~\cite{au2023ai}. High-stakes medical applications such as patient-facing diagnosis tools are especially vulnerable to such inaccurate information~\cite{sarraju2023appropriateness}. In response, adaptation methods in LLM for medicine have thrived, focusing on enhancing LLMs' domain-specific capabilities (Fig.~\ref{fig:overview}b). They include finetuning LLMs on medical data~\cite{singhal2023large}, adding relevant medical information to the prompts for LLMs via retrieval-augmented generation (RAG)~\cite{lewis2020retrieval,xiong-etal-2024-benchmarking}, and equipping LLMs with external tools to building AI agents achieving autonomous planning and task execution~\cite{nakano2021webgpt,jin2024agentmd}. With the increasing practice in developing LLM-based AI applications, it is becoming evident that cutting-edge performance is increasingly driven by the mixture of these adaptations~\cite{nori2023can} and systematic engineering of multiple AI components~\cite{compound-ai-blog}.

In this Perspective, we present an overview of adaptation techniques for developing LLM-based medical AI and the workflow to solidify the development (Fig.~\ref{fig:overview}). The adaptations can be concluded as:
\begin{itemize}[leftmargin=*]
    \item {\bf Model development} focuses on designing model architectures and employing learning algorithms to adapt model parameters for medical tasks. Techniques include injecting medical knowledge into general-purpose models through continual pretraining on medical datasets~\cite{Lin2024-wj} and fine-tuning, which aligns the model's outputs with domain-specific knowledge and human preferences~\cite{wang2024towards}.
    \item {\bf Model optimization} enhances model performance by optimizing its associated components, such as optimizing the input prompts~\cite{nori2023can} and implementing retrievers accessing external data to enable retrieval-augmented generation~\cite{lewis2020retrieval}.
    \item {\bf System engineering} enhances medical AI performance by breaking down tasks into well-defined, narrow-scope components. LLMs can serve as the computational core for each specialized component, which can be linked together to support complex workflows~\cite{khattab2023dspy}, or interact autonomously with other components to form agent-based systems~\cite{boiko2023autonomous}.
\end{itemize}

\highlight{Next, we provide actionable guidelines on when and how to adopt these adaptation methods based on the specific task parameters, e.g., time and cost constraints.} Additionally, we present concrete use cases that demonstrate the practical value of this framework. Finally, we discuss the associated challenges and opportunities for advancing LLM-based medical AI applications.

\section{Adapting large language models for medical AI}\label{sec:method}
This section will describe different strategies for adapting LLMs to build medical AI applications. 

\subsection{Model development: building medical-specific LLMs}

Generic LLMs such as ChatGPT benefit from vast parameters and are trained on vast, diverse datasets to develop a broad understanding of language~\cite{brown2020language}. A continual pretraining of generic LLMs on medical data, such as medical publications, clinical notes, and electronic health records (EHRs), can enhance LLMs' alignment in medical languages. For example, MEDITRON is based on the open-source LLaMA model~\cite{touvron2023llama}, further trained on medical data including PubMed articles and medical guidelines. It performs comparably to larger generalist models such as GPT-4~\cite{chen2023meditron} in medical question answering. Another model, PANACEA, shows the benefit of LLMs in clinical trial search, design, and patient recruitment via a continual pretraining of generic LLMs on clinical trial documents, including trial protocols and trial-related publications~\cite{Lin2024-wj}. \highlight{In addition, scaling model size using a mixture-of-experts strategy, where multiple LLMs (i.e., experts) are combined with a router network to select the appropriate expert during inference, has been found to yield superior performance at significantly lower computational cost compared to using a single, larger LLM~\cite{jiang2024mixtral}.}

Instead of {\it pretaining} with medical data without specific task supervision, an LLM can also learn from paired custom queries and their expected responses to perform multiple target tasks. A widely used technique is {\it instruction tuning}~\cite{wei2021finetuned}. For example, when PaLM, a 540-billion parameter LLM, underwent instruction tuning, it resulted in MedPaLM, which achieved 67.6\% accuracy on US Medical Licensing Exam-style questions~\cite{singhal2023large}. \highlight{When the computational resources are limited, we can also update a subset of the model's parameters or attach an additional small but learnable component to the model (e.g., prefix-tuning~\cite{li2021prefix} and LoRA~\cite{hulora2023}).}

Another important finetuning objective is alignment, which ensures LLM outputs are consistent with human preferences with high quality and safety. One prominent method for alignment is Reinforcement Learning from Human Feedback (RLHF), which was instrumental in developing ChatGPT~\cite{ouyang2022training}. RLHF starts with training a reward model based on human feedback, which then steers the LLM toward responses that align with human values and expectations using reinforcement learning.  An example of a medical application of alignment is LLaMA-Clinic, where multiple clinicians provide feedback to guide the models for generating high-quality clinical notes~\cite{wang2024towards}.

\subsection{Model optimization: strategies for improving LLM performance}

Prompt indicates the inputs to LLMs, which can include the task description, the expected input and output formats, and some input/output examples.  For example, this input ``Your task is to summarize the input clinical note. Please adhere to these summarization standards: [...list of criteria]. Refer to these examples: [...list of examples]. Keep in mind: [...list of important hints]," can guide LLMs towards generating the summary of patient notes. This practice leverages the principle of in-context learning, where LLMs adapt to new tasks based on the input prompts without additional training~\cite{brown2020language}. Research shows that the structure and content of the prompt significantly affect the model's performance. For instance, chain-of-thought prompting~\cite{wei2022chain} encourages LLMs to engage in multiple steps of reasoning and self-reflection, thereby enhancing the output quality. Another strategy involves ensembling~\cite{wang2022self,chen2024more}, where outputs derived from multiple prompts are synthesized to produce a final, more robust response.
In the medical domain, MedPrompt~\cite{nori2023can} has demonstrated its ability to outperform domain-tuned LLMs by combining multiple prompting techniques. Additionally, TrialGPT~\cite{jin2024matching} effectively adapts LLMs to match patient notes to the eligibility criteria of a clinical trial through prompting.

Handcrafted prompts are highly dependent on domain knowledge and trial-and-error, but they can still perform suboptimally and cause reliability issues in applications. Techniques such as automatic prompt generation~\cite{shin2020autoprompt} and optimization~\cite{cheng2023black} were proposed. These approaches can transform the adaptation of LLMs for medical tasks into a more structured machine-learning task. For example, requesting an LLM to summarize clinical notes could start with a simple prompt like ``Summarize the input clinical note." Using a collection of example notes and expert-written summaries, automatic summarization evaluation metrics can serve as the target for iterative prompt refinement. The prompt can ultimately evolve into a more advanced prompt consisting of a professional task description, representative examples, and a clear output format requirement description.

Retrieval-augmented generation (RAG) is an extension of prompting. It dynamically incorporates information retrieved from external databases into the model's inputs, supplementing LLM's internal knowledge~\cite{lewis2020retrieval}. In medicine, it is extensively employed to improve the factual accuracy of LLM responses to medical questions by fetching the evidence from medical literature, clinical guidelines, or medical ontology~\cite{xiong-etal-2024-benchmarking,wen2023mindmap}. \highlight{In practice, the efficacy of RAG relies on two aspects: (1) the quality of the external database, which should be high-quality and up-to-date~\cite {zakka2024almanac,arasteh2024radiorag} and (2) the performance of the retrieval systems, which are responsible for identifying and extracting the most relevant content to supplement the LLM's output~\cite{wang2022trial2vec, jin2023medcpt}.} 

\begin{figure}[htbp]
    \centering
    \includegraphics[width=\linewidth]{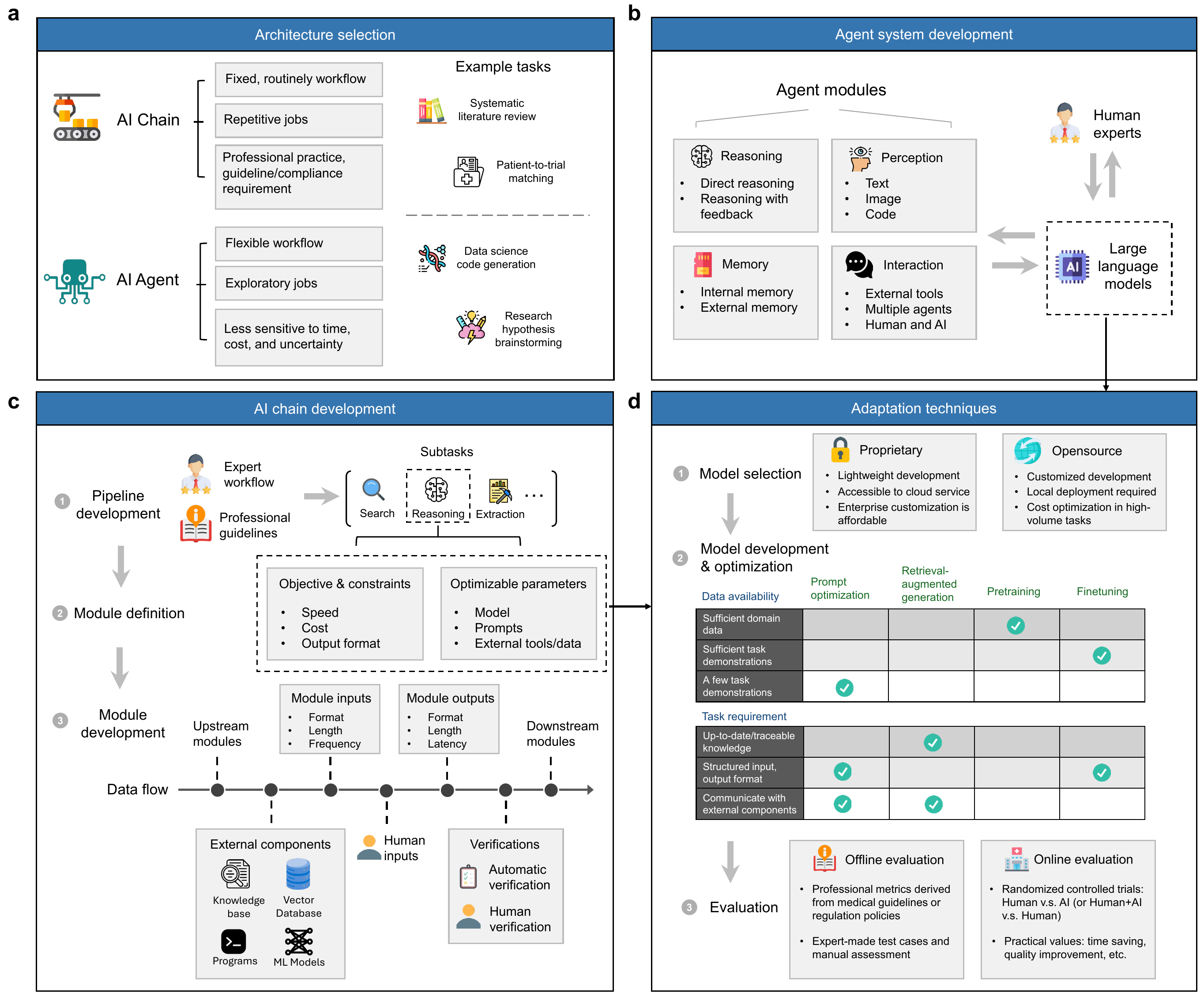}
    \caption{\highlight{This playbook outlines the process of adapting large language models (LLMs) for medical AI using a systems engineering approach.  (a) Selecting the overall architecture of the system should be based on the properties and requirements of the task at hand. (b) When building agent systems, four main modules need to be developed. LLMs act in different roles when equipped with different modules and interact with human experts to dynamically conduct the target task. (c) When building AI chain systems, we can first define the pipeline that decomposes the task into small steps following expert workflow or professional guidelines and then develop the module responsible for each step. (d) Adaptation techniques can be applied to enhance LLM's performance for the AI agent or for the AI module. Adaptation methods need to be selected according to data availability and task requirements.}}
    \label{fig:playbook}
\end{figure}

\subsection{System engineering: developing compound medical AI systems}
\highlight{Fig.~\ref{fig:playbook} illustrates the top-down system engineering process for adapting LLMs to medical AI applications. The process involves the following steps: (1) selecting the architecture to be developed (Fig.~\ref{fig:playbook}{a}); (2) designing the agent-based system (Fig.~\ref{fig:playbook}{b}) or the AI chain system (Fig.~\ref{fig:playbook}{c}); and (3) developing and fine-tuning LLMs to integrate with these systems (Fig.~\ref{fig:playbook}{d}).}

\highlight{
\paragraph{Architecture selection} The first step is to choose the overall system architecture: AI Chain~\cite{wu2022ai} and AI Agent~\cite{yao2022react}, depending on the nature of the tasks and requirements, as illustrated in Fig.~\ref{fig:playbook}{a}. AI Chain is suitable for fixed, routinely structured workflows, making it ideal for repetitive tasks, professional practices, and scenarios where adherence to guidelines or compliance is critical. Example applications include systematic literature reviews~\cite{wang2024acceleratingclinicalevidencesynthesis} and patient-to-trial matching~\cite{jin2024matching}, where processes are well-defined and consistent. On the other hand, AI agents are better suited for flexible workflows and exploratory tasks that are less sensitive to time, cost, and uncertainty. This paradigm is ideal for dynamic and creative tasks such as data science code generation~\cite{wang2024can} and brainstorming research hypotheses~\cite{openscholar}. The choice of architecture ensures that the system is aligned with the specific operational demands and objectives of the intended medical AI application.
}

\highlight{\paragraph{Agent system development} Developing AI agent systems boils down to designing and integrating key modules to facilitate the interactions between LLMs and human experts, as demonstrated in Fig.~\ref{fig:playbook}{b}.} AI agents embody a vision where AI systems actively solve problems through autonomous planning, knowledge acquisition using diverse tools, and iterative self-reflection~\cite{yao2022react}. This concept is supported by various agent frameworks that facilitate these capabilities, such as AutoGPT and AutoGen~\cite{autogpt, wu2023autogen}. Beyond external data sources, these agents can interact with various external tools, enhancing their operational scope. This includes interfaces for programming~\cite{tayebi2024large}, autonomous laboratory equipment~\cite{boiko2023autonomous}, and integration with other software programs~\cite{trinh2024solving, gu2024middleware}, broadening the potential applications of AI agents.

\highlight{
In general, an AI agent is composed of a base generalist AI like LLM and one or multiple modules so as to have different roles in conducting the task collaboratively with human experts. These modules include \textit{Reasoning}, which supports both direct reasoning and reasoning with feedback from humans, external tools, or other agents, to enhance decision-making capabilities; \textit{Perception}, which enables the agent to process and understand various data modalities such as text, images, and code; \textit{Memory}, comprising internal memory for storing temporary context (e.g., in prompts) and external memory for long-term information storage (e.g., in vector databases); and \textit{Interaction}, which supports communication with external tools, coordination with multiple agents, and collaboration between humans and AI. }

In medicine, AI agents are making remarkable progress in transforming healthcare delivery. For example, Polaris is an agent system designed to facilitate patient-facing, real-time healthcare conversations~\cite{mukherjee2024polaris}. It integrates a large language model (LLM) with automatic speech recognition for speech-to-text transcription and text-to-speech technology to convert LLM-generated text into audio, enabling seamless real-time interactions with patients. By leveraging agent-specific prompts and coordinating with specialized sub-AI models, such as those focused on medication management, laboratory and vital sign analysis, and clinical nutrition, Polaris functions as a multidisciplinary virtual specialist. Additionally, agents are increasingly used in biomedical research, assisting in complex data analysis and hypothesis generation~\cite{gao2024empowering}.

\highlight{\paragraph{AI chain development} For tasks that are usually fixed, repetitive, or strictly governed by professional guidelines, the AI chain is recommended to be developed (Fig.~\ref{fig:playbook}{c}). Rather than allowing LLMs to divide tasks and conduct them with several AI agents, experts can break down tasks into a chain of multiple steps.  This approach allows multiple LLM calls to create an AI chain, which enhances both the transparency and controllability of the final output~\cite{wu2022ai}.} An illustrative example is WikiChat~\cite{semnani2023wikichat}, which addresses hallucinations by sequentially employing LLMs to query Wikipedia, summarize, filter, and extract the retrieved facts, perform additional fact-checking, and finally draft responses with a round of self-reflection for refinement. As a result, WikiChat surpassed GPT-4 by 55\% in factual accuracy for question-answer tasks. In addition to optimizing performance, FrugalGPT~\cite{chen2023frugalgpt} optimizes query routing to reduce inference costs by using a cascade of LLMs with varying capabilities. This system utilizes weaker LLMs for initial processing and escalates to stronger LLMs only as necessary.

\highlight{AI chain development, as shown in Fig.~\ref{fig:playbook}{c}, consists of three primary stages: \textit{pipeline development}, \textit{module definition}, and \textit{module development}. The process begins with designing the pipeline based on expert workflows or professional guidelines, breaking it down into subtasks such as search, reasoning, and data extraction. Next, specialized modules are developed for each subtask. Each module is defined by its objectives (e.g., progress note generation) and constraints (e.g., response time and token usage) and is optimized using adjustable parameters such as the underlying models, prompts, and external tools or data sources. During module development, the focus is on the underlying model adaptations, considering the specifics of module inputs and outputs, including their format (e.g., text, image, code), length, and frequency (e.g., real-time or offline batch processing). For instance, the input length is an important factor for a module for summarizing multiple long documents, which requires an underlying model with a long context length optimized for improving fidelity to its inputs~\cite{tang2023evaluating}. For even longer inputs, advanced adaptations such as map-reduce, i.e., segmenting inputs, summarizing each segment, and consolidating those summaries into a global summary~\cite{bhaskar2023prompted}. This process also requires orchestrating data flows, integrating inputs from upstream modules, external components (e.g., knowledge bases, vector databases, or machine learning models), and human requests. The outputs need to undergo automatic or human verification to ensure accuracy and reliability. }

In medicine, AI chains have been developed to facilitate cohort extraction from EHR databases using natural language query~\cite{park2024criteria2query}. As generating correct SQL queries aligned with complex EHR database schema is challenging, the authors break down the task into three steps: concept extraction, SQL query generation, and reasoning, where the prompt for each step is designed and optimized separately. Chaining AI calls was also proved effective in parsing eligibility criteria into a structured format~\cite{datta2024autocriteria}. This method makes the first LLM call to extract the basic target entity information, then route to specialized extraction modules based on the entity type, where each prompt can be optimized, encoding expert knowledge. A post-processing module is attached to verify the outputs and aggregate the extraction results into a unified format.

\highlight{\paragraph{Adaptation techniques} 
The next step is the adaptation of LLMs, which includes two stages: \textit{model selection} and \textit{model development and optimization}, illustrated in Fig.~\ref{fig:playbook}{d}. In the model selection stage, the choice between proprietary and open-source models is guided by specific application requirements. Proprietary models, such as ChatGPT, are accessed via cloud services, enabling lightweight development and deployment without the need for local GPU resources but often at a higher cost, making them suitable for well-funded organizations. In contrast, open-source models such as finetuned LLaMa models offer greater flexibility for customization and are more cost-effective for high-volume tasks. They are particularly advantageous for local deployment scenarios due to privacy or security compliance requirements. }

\highlight{
The second stage emphasizes optimizing the selected model based on data availability and task requirements. Prompt optimization is the recommended initial approach for cases with limited or no labeled data, particularly when working with proprietary models~\cite{agrawal2022large}. However, if sufficient domain-specific data is available, continual pretraining can be applied to inject domain knowledge into generalist LLMs, enhancing their adaptability to specialized tasks. Nonetheless, sometimes the gain may be marginal if the domain-specific data are publicly available because training data for generic LLMs may already include the domain-specific data~\cite{jeong2024medical}. When prompt optimization fails to meet performance expectations, we recommend collecting more task demonstrations (i.e., input-output pairs) and finetuning the generalist models~\cite{zhang2024closing}. Finetuning offers several advantages: (1) it enables the model to handle tasks that are rarely encountered in the public corpus, and (2) it reduces the reliance on lengthy and complex prompts during inference, leading to faster processing times and lower costs~\cite{klang2024strategy}. 
}

\highlight{
Certain specialized tasks may require structured inputs and outputs. For instance, information extraction tasks or communication with physical devices require adherence to a specific input protocol and structured outputs, such as JSON objects, to ensure compatibility with machine parsing and processing~\cite{boiko2023autonomous}. Proprietary models often support features like JSON mode or function call capabilities~\cite{functioncall}, enabling such tasks to be handled efficiently through prompt optimization alone. Alternatively, open-source models can be fine-tuned on structured input-output pairs to achieve similar capabilities and can be further enhanced by employing constraint decoding frameworks~\cite{langchain}.} Additionally, for tasks requiring up-to-date knowledge, such as summarizing findings from the latest clinical trials, RAG is essential. It dynamically incorporates external knowledge into the LLM inference process, ensuring that outputs remain current and relevant. For instance, GeneGPT~\cite{jin2024genegpt} generates search queries to biomedical databases from the National Center for Biotechnology Information~\cite{sayers2019database} to enhance the factual accuracy of biomedical question answering, achieving an accuracy of 0.83 compared to ChatGPT's 0.12. 

\highlight{
\textit{Evaluation} is critical when adopting LLMs for medical applications, with the choice of evaluation metrics playing a central role. In many cases, general-domain metrics may not suffice. For example, in clinical note summarization tasks, general natural language processing metrics such as ROUGE~\cite{lin2004rouge} or BERTScore~\cite{zhangbertscore} may fail to adequately capture the quality of summaries in the medical domain. Instead, metrics tailored to the professional standards of medicine, such as completeness, correctness, and conciseness, should be prioritized~\cite{van2024adapted}. Designing these domain-specific metrics is often a collaborative effort between AI researchers and medical practitioners, and it is advisable to consult relevant medical guidelines or regulatory policies to ensure the chosen metrics are appropriate and comprehensive. Furthermore, real-world user studies serve as the gold standard for evaluation. Such studies typically involve two arms, one leveraging AI-assisted workflows and the other relying on manual efforts, to assess the practical value and effectiveness of medical AI systems~\cite{wan2024outpatient}.
}

\begin{table}[t]
  \centering
  \caption{A list of example use cases of adapting LLMs for medicine.}
\resizebox{\textwidth}{!}{
    \begin{tabular}{lllll}
    \toprule
    \textbf{Medical use case} & \textbf{Author, year} & \textbf{Adaptation methods} & \textbf{Data types} & \textbf{Model adopted} \bigstrut\\
    \midrule
    Outpatient reception & Wan, 2024\cite{wan2024outpatient} & Finetuning, Prompt optimization & Conversation & GPT-3.5 \bigstrut[t]\\
          & Habicht, 2024\cite{habicht2024closing} & Prompt optimization & Conversation & Unknown \\
    & Pais, 2024\cite{pais2024large} & Finetuning, Prompt optimization & Conversation & T5  \\
    Medical QA & Singhal, 2023\cite{singhal2023large} & Finetuning & Text  & PaLM \\
          & Nori, 2023\cite{nori2023can} & Prompt optimization & Text  & GPT-4 \\
          & Jin, 2024\cite{jin2024agentmd} & AI agent & Publications, Text & GPT-4 \\
    Multimodal medical QA & Jin, 2024\cite{Jin2024hidden} & Prompt optimization & X-rays & GPT-4v \\
          & Zhou, 2024\cite{zhou2024pre} & Finetuning & Skin images, Text & LLaMA, Vision transformer \\
    Radiology report generation & Zhang, 2024\cite{zhang2024generalist} & Finetuning & X-rays & Seq2Seq, Vision transformer \\
    Clinical note summarization & Van, 2024\cite{van2024adapted} & Prompt optimization, Finetuning & Clinical notes & GPT-3.5, GPT-4, LLaMA, T5 \\
    Clinical decision-making & Kresevic, 2024\cite{kresevic2024optimization} & Prompt optimization & Clinical guidelines & GPT-4 \\
          & Jiang, 2023\cite{jiang2023health} & Finetuning & EHRs  & BERT \\
          & Sandmann, 2024\cite{sandmann2024systematic} & Prompt optimization & Text  & GPT-3.5, GPT-4, LLaMA \\
    Patient-trial matching & Jin, 2023\cite{jin2024matching} & Prompt optimization, AI chain & Clinical notes, Clinical trials & GPT-4 \\
          & Park, 2024\cite{park2024criteria2query} & Prompt optimization, AI chain & Clinical notes, Clinical trials & GPT-4 \\
    Clinical research & Wang, 2024\cite{wang2024acceleratingclinicalevidencesynthesis} & Prompt optimization, AI chain & Publications, Clinical trials & GPT-4 \\
          & Tayebi, 2024\cite{tayebi2024large} & AI agent & Programs, Structured data & GPT-4 \\
          & Lin, 2024\cite{Lin2024-wj} & Finetuning & Clinical notes, Clinical trials & Mistral \\
    Information extraction & Keloth, 2024\cite{Keloth2024Advancing} & Finetuning & Text  & LLaMA \\
          & Huang, 2024\cite{huang2024critical} & Prompt optimization & Clinical notes & GPT-3.5 \\
    Drug discovery & He, 2024\cite{he2024novo} & Pretraining, Finetuning & Protein & PaLM \\
    Automatic medical coding & Wang, 2024\cite{wang2024drg} & Finetuning & EHRs  & LLaMA \bigstrut[b]\\
    \bottomrule
    \end{tabular}%
  }\label{tab:use_case_examples}%
\end{table}%

\begin{figure}[htbp]
    \centering
    \includegraphics[width=\linewidth]{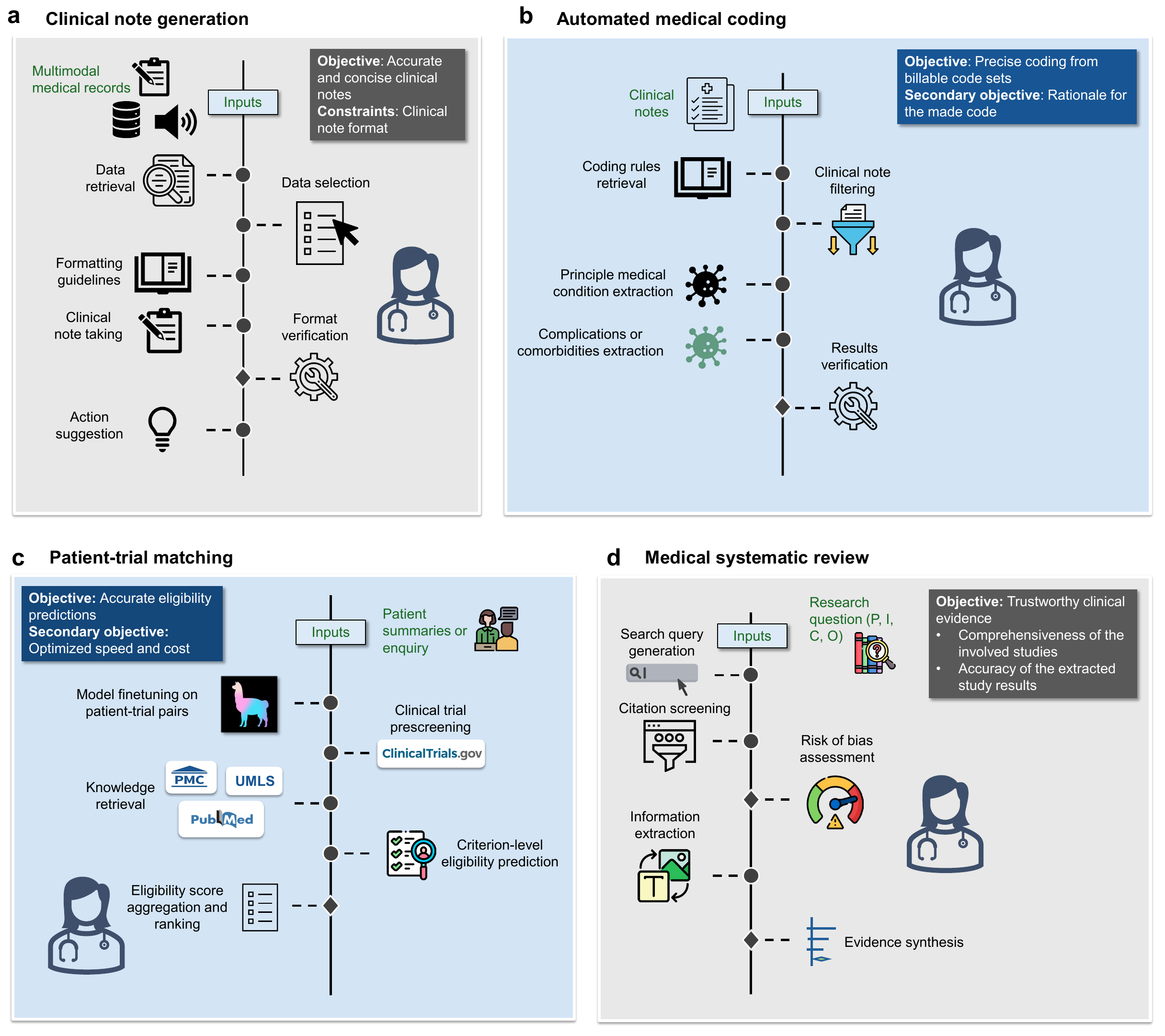}
    \caption{Illustration of example use cases adapting LLMs to medical AI tasks. \textbf{a,} AI chain crafted for clinical note generation, highlighting the expert involvement in selecting relevant patient data and adherence to external formatting guidelines. \textbf{b,} Automate medical coding can potentially benefit from an AI chain that employs two extraction modules designed for conditions and complications, respectively. \textbf{c,} A patient-trial matching pipeline adds a prescreening stage to reduce the candidate trial set. It also provides criterion-level assessment for users to select patients referring to various dimensions. \textbf{d,} Medical systematic review pipeline is built based on the established systematic review practice.}
    \label{fig:usecase}
\end{figure}

\section{Use cases of LLMs for medical AI applications}
Table~\ref{tab:use_case_examples} lists a number of studies that have explored adapting LLMs to medical AI tasks. In this section, we review a few example use cases, reinterpreting them through the lens of the proposed framework. Additionally, we discuss some potential improvements of those use cases.

\paragraph{Clinical note generation} Creating clinical notes is a routine task for physicians. These notes record the patient's medical history, present conditions, and treatment plans. Recently ambient documentation was introduced to record and transcribe provider-patient conversations and then ask LLMs to summarize them into a clinical note~\cite{abacha2023overview,Nelson2023-za}. This method is gaining popularity with the recent adoption of automated scribe software~\cite{yim2023aci}. However, the vanilla way of using LLMs is flawed in (1) Heterogeneity of note formats across specialties: The vanilla approach may not effectively adapt the note generation process to meet varying format preferences specific to different specialties (e.g., internal medicine versus general surgery), potentially leading to suboptimal documentation~\cite{wang2024towards}. (2) Lack of quality control: Ensuring the accuracy and completeness of generated notes is challenging due to the absence of accuracy measures and hallucination control for reference-free note generation. (3) Limited EMR integration: LLMs primarily accept text inputs of limited length. They may not fully integrate with electronic medical records (EMRs), often containing extensive and multimodal historical data. (4) Inadequate actionable insights: Unlike summarizing the inputs, LLMs usually perform unreliably in suggesting specific diagnostic tests or treatment actions following the generated summaries.

Next, we can provide a deeper technology analysis for developing such a specialized medical AI system for clinical note generation (Fig.~\ref{fig:usecase}a).
\begin{itemize}[leftmargin=*]
\item \textbf{Objective and constraints:} The primary objective is to produce accurate and concise notes that adhere to specific formatting requirements. This necessitates referencing ``best practice'' note formats of different specialties, outlining both the objectives and constraints for the output notes. Meanwhile, providing example inputs and outputs to LLMs' instructions can enhance clarity and accuracy. 

\item \textbf{RAG:} To customize note formatting according to specialty-specific needs, integrating a knowledge base into the pipeline is essential. This integration assists in pinpointing pertinent sections of clinical guidelines and dynamically retrieving closely related examples to enhance the instructions. Such adaptability makes the application more versatile across diverse clinical scenarios.

\item \textbf{Validation:} Human oversight is vital to validate the correctness of the generated notes. We can instruct LLMs to link each sentence in their summaries to a specific indexed part of the input, facilitating easy tracing of the source. Additionally, incorporating a layer of LLM self-reflection that compares the source and the target text can streamline this process. It provides clinicians with suggestions for omissions and inaccuracies in the summaries.

\item \textbf{EMR integration:} We can integrate LLMs with programming tools, such as an SQL interface to the underlying EMR database. This enhancement allows LLMs to dynamically access and review previous notes, only extracting necessary pieces to support the note generation process. In addition, this capability also enables the development of provider copilots to assist in the retrieval of necessary patient information at their fingertips in a natural language manner and empower them to focus more on care rather than system navigation.

\item \textbf{Action suggestion:} LLMs can surface the latest clinical guidelines via the help of an external knowledge base. For example, if the notes describe symptoms of hypertension, the LLMs can identify these symptoms and link them to the relevant part of the guidelines, suggesting appropriate diagnostic tests and medications.

\item \textbf{Task assistance}: AI agents can further assist providers in accomplishing specific tasks and reduce their administration workload. For example, follow-up emails are important to reinforce care plans, clarify medical instructions, and patient education, etc. LLM agents can create a follow-up email that includes a summary of the visit, test results, and next steps to ensure that patients understand their health information, adhere to treatment plans, and feel supported in their care journey. 

\item \textbf{Pipeline development:} We can structure the modules into a pipeline that facilitates collaboration between physicians and AI. In this pipeline, upon receiving patient data, the application initially displays the identified formatting guidelines (from the RAG module) and relevant segments of historical EMR data (from the EMR integration module) for the user to verify and select. Subsequently, refined instructions prompt the LLMs to generate the notes. This is followed by the validation module and the action suggestion module. A user-friendly interface will be designed to navigate users through this process.
\end{itemize}

\paragraph{Automated medical coding}
Medical billing codes, such as ICD-10, CPT, and DRG codes, are assigned by coding specialists through a time-consuming and manual process for healthcare payment systems. We can ask LLMs to translate EMR data into the most likely billing code~\cite{soroush2024large}. Fine-tuning LLMs on pairs of EMR data and billing codes can further enhance the performance~\cite{wang2024drg}. However, these methods are not yet satisfactory for production use because of (1) Inability to learn coding rules: Clinical coding goes beyond mere entity extraction. LLMs must be familiar with the potential list of billable codes relevant to specific input data. Moreover, the subtleties in the criteria for each billing code are crucial to avoid over-billing or under-billing. Compounding the complexity, each billing code system possesses its own set of nuanced rules and nomenclatures~\cite{topaz2013icd, dotson2013cpt, centers2023icd}. (2) Difficulty handling excessive EMR inputs: In inpatient settings, when a patient has a complex, multi-week hospital stay, it becomes problematic for an LLM to identify the most relevant sections of extensive medical record data without a sophisticated filtering strategy. (3) Lack of explainability: LLM-generated billing codes do not gain the trust of specialists without the rationale provided. We can approach these challenges considering (Fig.~\ref{fig:usecase}b):

\begin{itemize}[leftmargin=*]
\item \textbf{Objective and constraints:} The primary objective is to ensure precise coding from designated billable code sets, while the secondary objective is to provide a rationale. Both objectives require external knowledge of coding guidelines.
\item \textbf{RAG:} Medical coding guidelines are essential for accurately translating clinical notes into codes. To ensure compliance with these guidelines, coding rules must be integrated into the instructions for LLMs. This integration can be achieved through an RAG system, which enables the LLM to access relevant rules throughout the coding process. This method not only enhances the accuracy and relevance of the generated codes but also allows for the tracing back to the retrieved coding rules.
\item \textbf{EMR selection:} In cases of prolonged hospital stays, it is crucial to determine which notes should be processed by an LLM for billing code assignment. To optimize prediction performance, prioritizing the most clinically relevant notes, such as admission notes and discharge summaries, is recommended~\cite{wang2024drg}. Implementing a filtering strategy to remove duplicate content is advantageous for efficiency and cost-effectiveness. This addresses the common practice of copying and pasting in clinical documentation, which often results in redundant information for billing purposes.

\item \textbf{Coding-specific extraction modules:} The engineering of this extraction pipeline necessitates an advanced understanding of billing regulations. For instance, specific coding systems such as the Diagnosis-Related Group (DRG) used for hospital stays are structured hierarchically and hinge on two critical decision points: 1) establishing the primary medical issue as the diagnosis, and 2) assessing the presence of any complications or comorbidities~\cite{centers2023icd}. We can build two tailored extraction modules: one dedicated to accurately identifying the principal medical condition, and the other to detecting any associated complications or comorbidities. The final extraction results combine the two modules' outputs.

\item \textbf{Pipeline development:} The modules are organized into a streamlined pipeline that initiates with (1) retrieving relevant sections from the EMR; (2) allowing users to select the desired coding system; (3) retrieving coding rules from the corresponding guidelines; (4) directing the input to the appropriate coding-specific extraction module; and (5) generating the codes, with each code referencing the pertinent coding rules and EMR data for verification by a specialist. This methodical progression ensures that each stage of the coding process is accurate and efficiently managed, optimizing data integration for billing purposes.
\end{itemize} 

\paragraph{Patient-trial matching} Matching patients with appropriate clinical trials is crucial for ensuring sufficient trial recruitment and significantly impacting trial outcomes. The manual recruitment process, which involves screening surveys, is labor-intensive, time-consuming, and prone to errors. Efforts have been made to optimize prompts for powerful LLMs like GPT-4 to assess patient eligibility by matching patient records and trial eligibility criteria~\cite{wornow2024zero}. To mitigate concerns about Protected Health Information (PHI) leakage when using proprietary models, another approach has been proposed: leveraging the knowledge of these proprietary models to generate synthetic patient-trial pairs, which are then used to fine-tune smaller open-source LLMs for secure local deployment~\cite{nievas2024distilling}.  However, challenges remain, including (1) scalability: Existing methods typically assume a limited set of candidate trials or patients. However, with over 23,000 active trials on ClinicalTrials.gov and patient databases containing millions of records, screening on a patient-by-patient or trial-by-trial basis using LLMs becomes intractable. (2) Lack of medical expertise: Generalist LLMs may still struggle with medical reasoning when assessing eligibility. For example, a patient record noting a ``prescription history of blood thinners due to recurrent chest pain'' was incorrectly labeled as ``not enough information'' by GPT-4 in response to the criterion ``patient must have a history of cardiovascular disease.''~\cite{jin2024matching}. A system engineering can be made to approach this task (Fig.~\ref{fig:usecase}c).

\begin{itemize}[leftmargin=*]
\item \textbf{Objective and constraints:} The primary objective is to achieve accurate eligibility predictions, with secondary objectives focused on optimizing speed and cost, particularly when handling large datasets.
\item \textbf{Finetuning:} Enhancing the medical capabilities of LLMs through fine-tuning on high-quality patient-trial pairs is recommended~\cite{nievas2024distilling}. This necessitates a streamlined method for annotating patient eligibility against trial criteria or access to records of previous trials' participants.
\item \textbf{RAG:} Correlations between medical terms, such as therapies and conditions, can be referenced from clinical guidelines. Additionally, medical ontologies can map umbrella terms to their specific subsets by following the hierarchy of medical concepts. These resources can be integrated as external tools to enhance the LLM's reasoning capabilities for more accurate eligibility assessments~\cite{unlu2024retrieval}.
\item \textbf{Prescreening:} Scalability challenges can be mitigated by adding a prescreening module before invoking eligibility assessment. For example, TrialGPT~\cite{jin2024matching} introduces a trial retrieval step by generating search queries based on patient summaries, effectively reducing the candidate set by over 90\% without compromising the recall of target trials. Additionally, LLMs can perform information extraction on both patient records and eligibility criteria, followed by entity matching as a prescreening step~\cite{wong2023scaling}.
\item \textbf{Pipeline development:} Concluding the above adaptation methods, a pipeline may constitute: (1) fine-tuning underlying LLMs with patient-trial data; (2) initiating a prescreening process for clinical trials to narrow down the candidate pool based on the target patient; (3) extracting key terms from patient records and retrieving relevant knowledge from clinical guidelines and medical ontologies; (4) invoking LLM calls for the final eligibility assessment.
\end{itemize}

\paragraph{Medical systematic review} Clinical evidence can be synthesized through the review of medical literature, but this process is increasingly challenged by the rapid growth of published research. LLMs have been employed to synthesize evidence from multiple articles using text summarization techniques~\cite{shaib2023summarizing,tang2023evaluating}. However, this text summarization approach raises significant concerns about the quality of the outputs, including issues like lack of specificity, fabricated references, and potential for misleading information~\cite{yun2023appraising}. Moreover, conducting a systematic review involves more than just summarization; it encompasses multiple steps, as outlined in PRISMA~\cite{page2021prisma}. AI should be integrated into the established systematic review workflow with careful consideration of user experience optimization, such as offering verification, referencing, and human in the loop~\cite{wang2024acceleratingclinicalevidencesynthesis}. From a systems engineering perspective, the following elements should be incorporated (Fig.~\ref{fig:usecase}d).

\begin{itemize}[leftmargin=*]
    \item \textbf{Objective and constraints:} The primary objective is to generate trustworthy clinical evidence from medical literature, with sub-objectives including the comprehensiveness of collected studies and the accuracy of extracted results. Strict formatting constraints should be established for outputs at each step. For example, synthesized evidence must be directly supported by the results of individual studies, ideally presented in the form of a meta-analysis.
    \item \textbf{Search query generation:}  Generating search queries requires a deep understanding of both medicine and library sciences. Prompt optimization can be employed for LLMs to enhance their ability to generate comprehensive search queries~\cite{wang2023can}.
    \item \textbf{Citation screening:} After identifying thousands of citations, a refined screening process is essential to ensure relevance to the research question, covering the population, intervention, comparison, and outcome (PICO) elements. Additionally, a risk-of-bias assessment should be conducted to filter out low-quality citations. Developing specialized prompts that incorporate itemized scoring guidelines can enhance the accuracy and consistency of this screening process.
    \item \textbf{Information extraction:} A PDF extraction module is required, based on Optical Character Recognition (OCR) techniques, to extract the original content, including tables and figures, from the studies. Additionally, the LLM outputs should include references to the original content, enabling users to verify and correct any potential misinformation.
    \item \textbf{Evidence synthesis:} Extracting numerical results, particularly from clinical studies, can be challenging and requires accurate extraction of evaluated endpoints, sample sizes, and treatment effects. To enhance the accuracy of this process, LLMs can be equipped with an external program interface that assists in numerical reasoning during evidence extraction and synthesis~\cite{wang2024acceleratingclinicalevidencesynthesis}.
\end{itemize}

\paragraph{Mapping data privacy legislations} The challenge of aligning diverse and evolving privacy, data protection and AI legislation with standardized frameworks such as the National Institute of Standards and Technology (NIST) has become increasingly complex, especially when dealing with multiple jurisdictions~\cite{National-Institute-of-Standards-and-Technology2020-dy,National-Institute-of-Standards-and-Technology2024-em}. For instance, in a recent U.S.-based project for a health application, an LLM pipeline is implemented by IQVIA to analyze privacy legislation across several states, identifying over 3,000 overlaps between legislative requirements and NIST privacy actions~\cite{Covert2020-py}. This level of analysis, achieved in a short timeframe, would have been near impossible to accomplish manually at the necessary scale and depth. The same approach is being used to identify controls and governance for a data platform in the Middle East.

The LLM pipeline we devised automates identifying and mapping privacy, cybersecurity, and AI risk management actions, offering a scalable solution that can navigate a complex and rapidly changing regulatory landscape. The pipeline enables more efficient compliance management by processing extensive legal texts and implementation guidelines and aligning them with structured frameworks. Integrating a human-in-the-loop, expert verification and tuning process, while leveraging established NIST crosswalks to validate accuracy, provides a robust method for tackling large-scale data and AI compliance challenges~\cite{Tabassi2023-to}.

\begin{itemize}[leftmargin=*]

\item	\textbf{Objective and constraints}: The primary objective is to automate the mapping of regional and global legislation to the NIST Frameworks, providing a scalable method for compliance management. The secondary objectives include ensuring the comprehensiveness of the legislative documents analyzed, the accuracy of identified compliance actions, and the risk-based prioritization of implementation activities. 

\item	\textbf{Data collection and preprocessing}: The process begins by collecting relevant legislative and regulatory texts, including regional and sectorial guidelines. This data is ingested into the LLM pipeline, which is designed to process large volumes of documents efficiently. Preprocessing includes document parsing, natural language normalization, and segmentation, allowing the model to focus on actionable items relevant to compliance management.

\item \textbf{Risk-based thresholding}: Once the LLM generates its initial matches against the frameworks, a scoring and thresholding system is applied. Each identified compliance action is assigned a score based on its relevance and alignment with the NIST compliance actions. A risk-based assessment is conducted by scoring overlap with the sourced documents, weighted on the importance of the business context (e.g., usability, implementation guardrails, effort). 

\item \textbf{Human-in-the-Loop expert guidance}: While the LLM provides an initial mapping and scoring, experts validate the accuracy and relevance of the overlaps, review and fine-tune the LLM's outputs, resolving any ambiguities or jurisdiction- and sectorial-specific variations that the model may misinterpret. Experts also interpret and summarize results based on client needs while providing evidence-based decision-making from the mapping and scoring.

\item \textbf{Benchmarking and confidence}: To ensure accuracy, the pipeline incorporates a validation step using pre-established NIST crosswalks developed by legal and other experts. These crosswalks serve as benchmarks, allowing the system to compare its outputs against known mappings of specific legislation to the NIST Frameworks. The LLM pipeline also uses advanced confidence elicitation strategies to mitigate LLM overconfidence in ambiguous or novel medical situations. 

\item \textbf{Evidence synthesis and validation}:  For validation, the LLM pipeline's outputs are synthesized into a final report, which includes detailed references to the relevant legislative texts and framework categories. The synthesis process ensures traceability, allowing users to easily verify the source of each mapping decision, as evidence of how features are prioritized.

\end{itemize}

\section{Opportunities and challenges in LLMs for medical AI}
Here, we describe the challenges that need to be addressed to embody the benefits of adapting LLMs to medical AI, with a focus on three critical areas: enhancing multimodal capabilities, ensuring trustworthiness and compliance, and managing system lifecycle through evaluation and continuous optimization.

\highlight{\subsection{Multimodality}}
Multimodal capabilities represent a key growth direction for LLMs in medical applications. A patient journey naturally comprises many information-rich modalities such as lab tests, imaging, genomics, etc. While generalist AI has demonstrated amazing capabilities in understanding and reasoning with biomedical text (e.g., MedPrompt~\cite{Nori2023-cd}), competence gaps abound in the multimodal space, with vision-language model being a prominent example. E.g., GPT-4V performs poorly on identifying key findings from chest X-rays, even compared to much smaller domain-specific models (e.g., LLaVA-Rad~\cite{Chaves2024-ro}). Efficiently bridging such multimodal competency gaps thus represents a key growth frontier for medical AI. Progress is particularly fast in biomedical imaging, from harnessing public image-text data (e.g., BiomedCLIP~\cite{zhang2023large-scale}, PLIP~\cite{Huang2023-mw}, CONCH~\cite{Lu2024-lv}) to efficiently training vision-language models (e.g., LLaVA-Med~\cite{Li2023-vj}) to learn text-guided image generation and segmentation (e.g., BiomedJourney~\cite{Gu2023-li}, RoentGen~\cite{Bluethgen2024-ex}, BiomedParse). While promising, challenges abound in multimodal medical AI, from pretraining in challenging modalities (e.g., GigaPath~\cite{Xu2024-hq}) to multimodal reasoning for precision health.

\highlight{\subsection{Trustworthiness and compliance}}
\highlight{While adapting powerful AI models like LLMs to medicine holds great promise, significant challenges persist in building trustworthiness and ensuring compliance with their use. Here, we identified several challenges: (1) hallucinations, (2) privacy risks, (3) explainability, and (4) regulations.}

The risk of ``hallucinations'' in LLM outputs, where the model generates plausible but incorrect or fabricated information, is widely mentioned~\cite{huang2023survey}. In the medical domain, such errors can have severe consequences, including misdiagnoses, inappropriate treatments, and flawed research conclusions~\cite{thirunavukarasu2023large}. To address hallucinations, retrieval-augmented generation (RAG) is often employed to guide LLM outputs and ground them in verifiable citations~\cite{gao2023enabling}. However, failure modes persist, such as LLMs citing incorrect sources or hallucinating citations altogether~\cite{wu2024well}. Future efforts should focus on fine-tuning models with high-quality, domain-specific datasets, implementing rigorous validation mechanisms, and incorporating human-in-the-loop workflows~\cite{SAMS2024}, where medical professionals review and correct AI-generated content to ensure accuracy and reliability.

The development and deployment of medical AI applications involve multiple stakeholders, including medical data providers (such as health systems, pharmaceutical companies, and real-world data companies), model providers (like healthcare AI startups), and users (comprising clinicians, patients, and these data-providing companies). AI developers play a crucial role in constructing pipelines that process data from these data providers into actionable knowledge, which is then made accessible to users by leveraging intelligence from model providers. It has been noted that LLMs are vulnerable to attacks using malicious prompts that aim to extract their training data~\cite{carlini2021extracting}. \highlight{This risk necessitates the deidentification of the PHI information from training data. When now real data can be acquired, synthetic training data can be incorporated~\cite{theodorou2023synthesize,das2023twin}, which can be further protected with differential privacy~\cite{torfi2022differentially}.} Moreover, when AI developers handle data from multiple providers and present it to various users, it is essential to implement access controls within the AI applications. These controls ensure that users can only access specific data sets relevant to their queries. In addition, protecting user inputs is crucial, especially if they are utilized to optimize the pipeline and stored in prompts or databases for RAG. LLMs must prevent inadvertently disclosing one user's data to another, maintaining strict confidentiality in user interactions.

\highlight{LLM's interpretability remains a critical challenge. It appears to show ``emergent" intelligence but that property may also disappear when we take different evaluation metrics~\cite{schaeffer2024emergent}.} As deep learning models, LLMs often function as black boxes, making it difficult to trace the reasoning behind specific outputs. This lack of transparency can hinder clinical decision-making and raise concerns about accountability. To address these issues, developing explainable AI methodologies is essential for gaining a deeper understanding of LLMs~\cite{zini2022explainability}. At present, techniques such as chain-of-thought and program-of-thought can be employed to reveal the step-by-step reasoning and operational processes behind LLM outputs, improving their interpretability in medical applications~\cite{wei2022chain,chen2023program}. \highlight{Researchers have also tried to dive into the inner workings of AI systems by introducing mechanistic interpretability, which is promising to provide a more interpretable and controlled LLM behavior~\cite{bereska2024mechanistic}.}

The development, deployment, and production of LLMs in medical practice must adhere to strict regulatory standards and compliance requirements to ensure patient safety and data privacy~\cite{li2017exploration}. Regulatory bodies such as the FDA, EMA, and others have established guidelines for medical devices, which increasingly include AI-driven tools~\cite{wu2023characterizing}. Developers of medical AI applications must navigate these regulations, establish robust practices, and process procedures to ensure that their models are validated, transparent, and compliant with relevant laws, such as HIPAA in the U.S. Additionally, as AI applications are continuously updated and retrained, maintaining compliance over time requires ongoing monitoring, documentation, and potentially re-certification.

\highlight{\subsection{Evaluation and continuous improvement}}
A medical AI system may comprise various modules and pipelines, each of which has the potential to malfunction, posing challenges in output assessment and debugging. Human oversight can be integrated to enhance validation, allowing users to confirm the accuracy of outputs through the provided references~\cite{bakken2023ai}. Furthermore, validation processes can be automated by leveraging LLM capabilities. For example, assertions can be embedded within the pipeline, enabling LLMs to self-correct their outputs during inference~\cite{singhvi2023dspy}. The increasing complexity of LLM-based systems, consisting of multiple interconnected components, can overwhelm any individual's capacity to manage the entire architecture. LangSmith~\cite{langsmith} is a DevOps tool for AI applications that aims to support deploying and monitoring LLM-powered applications. It allows developers to visualize traces and debug problematic components. Additionally, it facilitates the collection and extraction of erroneous inputs and outputs, which can then be utilized to build and augment validation datasets to enhance the applications.

Maintaining medical AI applications poses significant challenges in ensuring system stability and reliability. One major concern is the variability of the underlying LLMs, as version updates can alter model behavior and capabilities, potentially disrupting the functionality of integrated systems~\cite{chen2023chatgpt}. Additionally, the interdependence of system components means that changes to one element can impact the overall performance, necessitating rigorous testing and calibration whenever modifications are made. Furthermore, managing distribution shifts, particularly in edge cases, remains a critical issue.  An effective strategy to address these challenges is the construction of robust development datasets for evaluating and fine-tuning AI pipelines. These datasets can be sourced from real-world tasks or simulated scenarios generated by LLMs~\cite{deepeval}. However, current approaches often rely on heuristic methods, underscoring the need for further research to enhance dataset creation. \highlight{Another significant challenge lies in selecting appropriate evaluation metrics, as many standards are embedded in specialized medical guidelines or regulatory documents. Bridging the knowledge gap between AI scientists and medical practitioners through increased collaboration and interdisciplinary research is essential for establishing reliable evaluation frameworks.}

User feedback is a crucial source of supervision for the continual optimization of AI pipelines. This feedback can take various forms, ranging from explicit expressions of preference, such as likes or dislikes, to more subtle indicators found in user interaction logs or direct textual comments on specific aspects of the system. Such feedback is valuable not only for refining the pipeline's final outputs but also for improving any intermediate stages where user engagement occurs, whether through active participation or the generation of usage logs. Techniques like Reinforcement Learning from Human Feedback (RLHF)~\cite{ouyang2022training} have been used to enhance LLM models based on ranked human preferences. Recently, a framework called TextGrad~\cite{yuksekgonul2024textgrad} has shown promising results by enabling the backpropagation of textual feedback within LLM pipelines to optimize prompts across different components.


\section{Conclusion}
Generalist AI models have the potential to revolutionize the medical field. A range of adaptation methods have been proposed to tailor these models for specialized applications. In this Perspective, we documented existing adaptation strategies and organized them within a framework designed to optimize the performance of medical AI applications from a systems engineering perspective. Our analysis and discussion of published use cases demonstrate the benefits of this framework as a systematic approach to developing and optimizing LLM-based medical AI. However, we also recognize the potential challenges that arise as the complexity of medical AI applications increases, particularly in monitoring, validation, and maintenance. Future research and development are essential to solidify the utility of LLM-driven medical AI applications, enhance patient outcomes, democratize access to quality healthcare, and reduce the workload on medical professionals.

\clearpage

\captionsetup[figure]{name=Extended Fig.}
\captionsetup[table]{name=Extended Table}

\begin{appendices}

\end{appendices}

\clearpage






\bibliographystyle{naturemag}
\bibliography{main}

\end{document}